\documentclass[letterpaper, 10 pt, conference]{ieeeconf}  

\IEEEoverridecommandlockouts                              

\usepackage{amsmath} 

\usepackage{cite}
\usepackage{amsmath,amssymb,amsfonts}
\usepackage{algorithmic}
\usepackage{graphicx}
\usepackage{sidecap}
\usepackage{textcomp}
\usepackage{lipsum}  
\usepackage{xcolor}
\usepackage[font=footnotesize]{caption}
\usepackage{setspace}
\usepackage{subfig}
\usepackage{tabularx, booktabs}

\begin{document}

\title{\LARGE \bf The Coupling Effect: Experimental Validation of the Fusion of Fossen and Featherstone to Simulate UVMS Dynamics in Julia  \\

\thanks{Collaborative Robotics and Intelligent Systems (CoRIS) Institute, Oregon State University, Corvallis OR 97331, USA {\tt\footnotesize \{kolanoh, palmeeva, joseph.davidson\}@oregonstate.edu}}

\author{Hannah Kolano, Evan Palmer, Joseph R. Davidson}
}%


\maketitle

\begin{abstract}
As Underwater Vehicle Manipulator Systems (UVMSs) have gotten smaller and lighter over the past years, it is becoming increasingly important to consider the coupling forces between the manipulator and the vehicle when planning and controlling the system. A number of different models have been proposed, each using different rigid body dynamics or hydrodynamics algorithms, or purporting to consider different dynamic effects on the system, but most go without experimental validation of the full model, and in particular, of the coupling effect between the two systems. In this work, we return to a model combining Featherstone’s rigid body dynamics algorithms with Fossen’s equations for underwater dynamics by using the Julia package RigidBodyDynamics.jl. We compare the simulation's output with experimental results from pool trials with a ten degree of freedom UVMS that integrates a Reach Alpha manipulator with a BlueROV2. We validate the model's usefulness and identify its strengths and weaknesses in studying the dynamic coupling effect.
\end{abstract}

\color{black}
\section{Introduction}


According to a recent study, man-made structures covered 32,000 km$^2$ of the ocean floor in 2018, projected to increase to 39,400 km$^2$ by 2028~\cite{Bugnot2020}. These structures have to be routinely inspected and maintained. It is safer to deploy Underwater Vehicle Manipulator Systems (UVMSs) for these tasks than human divers. Additionally, whereas large work-class UVMSs are expensive to deploy, smaller inspection class vehicles are cheaper, lighter, and more accessible, making them ideal to deploy on a wider scale. 

\begin{figure}[t]
	\centering
\includegraphics[width =\columnwidth]{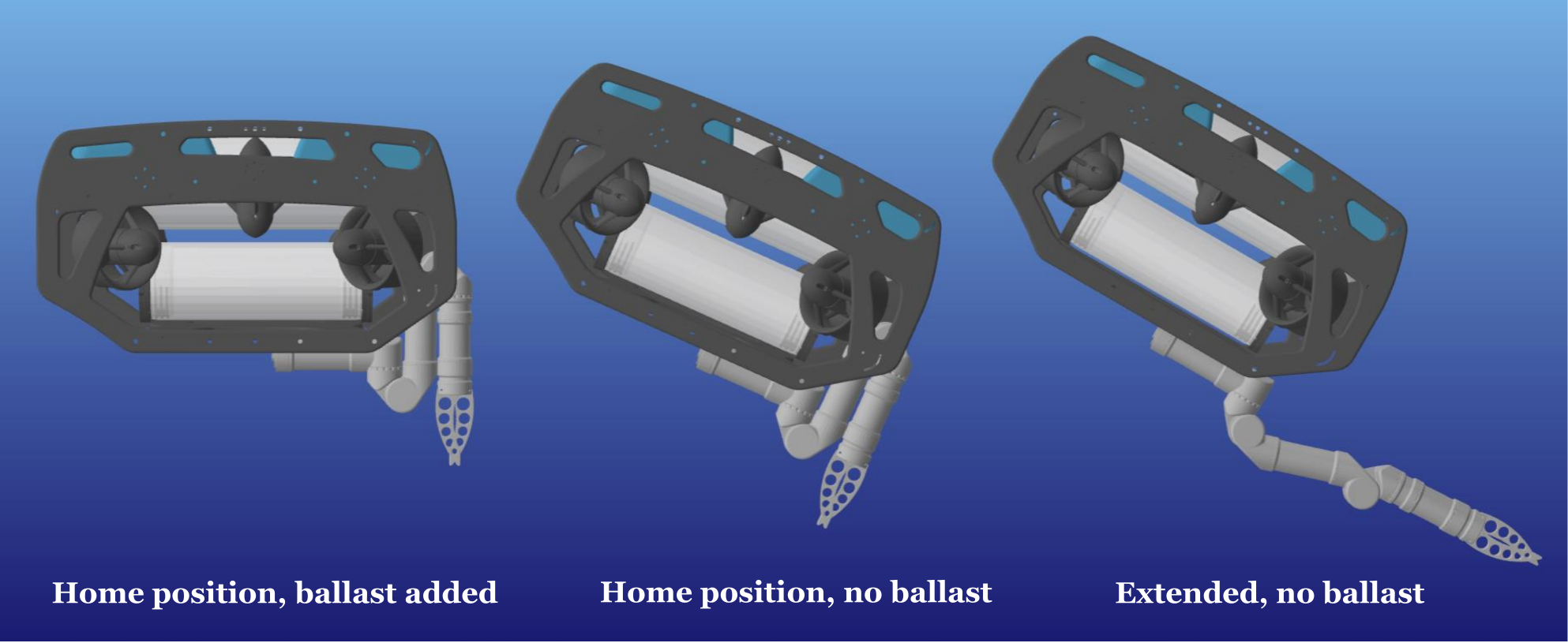}
\caption{A lightweight UVMS. (\textit{Left}) The equilibrium position if ballast is added to offset the manipulator weight. If no ballast is added, the vehicle pitches significantly at the home position (\textit{center}) and even more in the extended position (\textit{right}).}
\label{fig:abstract}
    \vspace{-15pt}
\end{figure}

However, lightweight UVMSs face challenges that larger vehicles do not. In particular, the relatively large mass ratio between the manipulator and the vehicle causes reactive motions of the vehicle when the manipulator is actuated. Smaller vehicles tend not to have active control of roll and pitch, so they are unable to compensate for the coupling forces with opposing forces. In these cases, knowing the extent to which the manipulator motion will change the orientation of the vehicle is particularly important, especially during tasks requiring precision manipulation (Figure \ref{fig:abstract}).


 Not a lot of work has been done to study the kinematic impact of manipulator motion on vehicle response for underactuated vehicles. In particular, simulators that combine stable rigid body dynamics of open chains with hydrodynamics are sparse. Many models have been proposed in recent years, but there is a lack of experimental validation of the models themselves in realistic conditions. As operating UVMSs at sea is expensive and high-risk, simulations could be very useful for generating data, training learning-based controllers, etc. However, crossing the sim-to-real gap requires a simulation environment that faithfully captures some of the complex physics of the real world. In practice, implementations in the real world tend to rely on proxy solutions, such as constraining manipulator movement to a box \cite{Lillo2019}, but minimizing manipulator motion in this way unnecessarily limits the operational workspace and constrains capabilities.  

In this paper, we study UVMS dynamic coupling in depth. To complete this analysis, we developed a UVMS simulator that combines Fossen’s hydrodynamics equations with Featherstone’s open chain rigid body dynamics algorithms through the use of Julia’s RigidBodyDynamics.jl package, described in Section \ref{sec:Methodology}. We completed a number of manipulator trajectories in the simulation with an unactuated vehicle. We then deployed a UVMS at a wave research lab and executed the same trajectories, using underwater motion capture to compare the simulation's predictions with the real physical system (Section~\ref{sec:experiments}). \textit{Our primary contribution in this paper is experimental validation of a dynamic modeling framework that has been widely adopted in underwater robotics}. Our results (Section~\ref{sec:Results}) indicate that while the dynamic model captures much of the system's qualitative behavior, there is error from modeling assumptions and simplifications, and additional work is needed to better replicate the complex physics of the underwater environment.

\section{Related Work}
\label{sec:RelatedWork}

\subsection{Dynamic Coupling Effects}
When a manipulator is attached to an underwater vehicle, forces are transmitted between the two systems, an effect called \textit{coupling}. Traditionally, UVMSs have large vehicle-to-manipulator mass ratios (e.g. work class systems), so coupling can usually be ignored. However, as the vehicle-to-manipulator mass ratio decreases, methods that handle the dynamic coupling effect become critical. Schempf and Yoerger \cite{Schempf1992} were one of the first to recognize the adverse effects of ignoring coupling forces. The authors demonstrated that even a 20:1 mass ratio between the vehicle and the manipulator caused significant unwanted coupling effects. Other researchers later confirmed that modeling the hydrodynamic effects was essential to station keeping and end effector accuracy \cite{Wang1995,Rigaud1998}. 

The coupling effect has become a prescient problem in recent years as underwater vehicles have become lighter and more accessible. In particular, Barbalata et. al extensively explored the coupling effect on a lightweight UVMS in simulation \cite{Barbalata2015, Barbalata2018}. Because the authors used a computed torque control scheme, they were able to calculate the coupling forces and compensate for them with the controller. Similarly, the Ocean One UVMS uses a whole-body computed torque controller to account for coupling effects on their custom-built system \cite{Brantner2020}. 

Most off-the-shelf, commercial underwater manipulators do not provide low-level torque control at each actuator, which precludes the use of computed torque controllers. Therefore, in practice, most UVMS controllers are kinematics based, such as SAUVIM \cite{Marani2009}, TRIDENT \cite{Sanz2013}, MARIS \cite{Simetti2015}, and ROBUST \cite{Dai2020}. Recently, the DexROV project showed promising whole-body kinematic control results in a kinematics-only simulation. However, during sea trials the vehicle was clamped to an underwater structure, negating the need to understand the coupling effect \cite{DiLillo2021}. Kinematically, the manipulator was constrained to a virtual box to prevent self-collisions and over-extension \cite{Lillo2019}. 

More recently, efforts have been made to model the coupling effect. Xiong et al.~\cite{uvdms} developed a metric to describe the coupling effect and used it to characterize the impact of different design factors on the magnitude of coupling. The authors found that in highly coupled systems the orientation of the vehicle was most affected. However, the model was not validated on hardware, so its utility is unknown. Huang et al.~\cite{positionuvdms} also discuss metrics to characterize the degree of coupling, but only with static positions, not dynamic motions. A quasi-Lagrange formulation was used in \cite{jointflex} to model a UVMS and its coupling effect. The authors also considered non-stiff joints, modeling them as torsion springs, and included vortex-induced vibrations into their model. The authors did their modeling on a 2-link manipulator in simulation, a simplification of the kinematically redundant systems deployed in the field. 

\subsection{Underwater Simulators}
A significant hurdle was finding a simulator capable of accurately modeling the coupling forces in underwater environments. Many simulators exist for rigid body dynamics, and others exist for hydrodynamics, but very few exist that combine both. While a more general review of underwater simulators can be found in \cite{simreview}, the work does not discuss the rigid body dynamics capabilities of the simulators. One of the most mature programs for UVMSs is UUV Sim \cite{UUVSim}; however, since its backend uses Gazebo, the way UUV Sim handles added masses can create numerical instabilities, as described in Section \ref{sec:sim}. After a preliminary exploration of our model (Sec.~\ref{sec:sim}) in UUV Sim, we determined that it was too unstable to accurately capture the coupling effect and provide reliable forces. 

Other well-known underwater simulators include UWSim \cite{UWSim} and HoloOcean \cite{HoloOcean}. UWSim has complete hydrodynamics, but does not have the capability to handle chains of rigid bodies. A more recent simulator, HoloOcean, is built on Unreal Engine 4, which has opaque physics primarily meant to create realistic graphics and not accurate forces. HoloOcean itself is built mostly for multiagent applications as well as modeling sonar sensors. 

The most promising simulator for examining the coupling effect is the Stonefish Simulator, built by Cie\'slak \cite{stonefish}. Stonefish is built upon the Bullet physics engine \cite{bullet} and uses geometry-based hydrodynamics. We use similar hydrodynamic assumptions and rigid body dynamics implementations here. For this work, we developed our own simulation in Julia because we desired more transparency and control over the underlying dynamics, but our validation should hold true for Stonefish as well.

\section{Methodology}
\label{sec:Methodology}

Our work focuses on dynamic coupling and how a rigid body model commonly adopted in underwater robotics compares with real world data. We start by describing the rigid body model of a ten degree of freedom (DOF) system --- and its underlying assumptions --- and the simulation environment. For this paper, we assume a calm hydrodynamic environment and leave the study of energetic waves, currents, etc. to future work. To obtain real world data of UVMS states during manipulator motion, we deployed the UVMS at a wave basin and used an underwater motion capture system to record UVMS position and orientation during execution of manipulator trajectories.

\begin{figure}[]
\vspace{5pt}
	\centering
\includegraphics[width =\columnwidth]{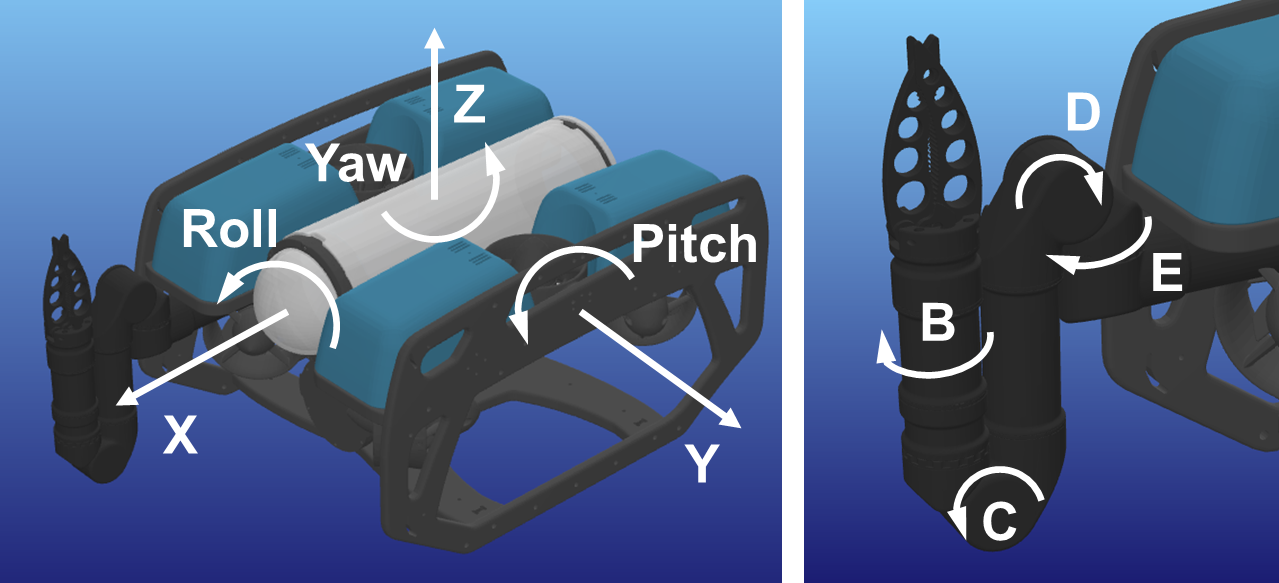}
\caption{Images of the BlueROV2/Alpha UVMS, as visualized in MeshCat from the Julia simulation. \textit{(Left)} A right-handed Cartesian coordinate frame is assigned at the vehicle's center of mass, using forward-left-up notation. \textit{(Right)} The Alpha's four joints: a base joint parallel to yaw, two joints to extend the arm, and a rotational joint at the wrist.}
\label{fig:UVMS}
    \vspace{-10pt}
\end{figure}

\begin{figure*}[]
\centering
\vspace{5pt}
\includegraphics[width =\linewidth]{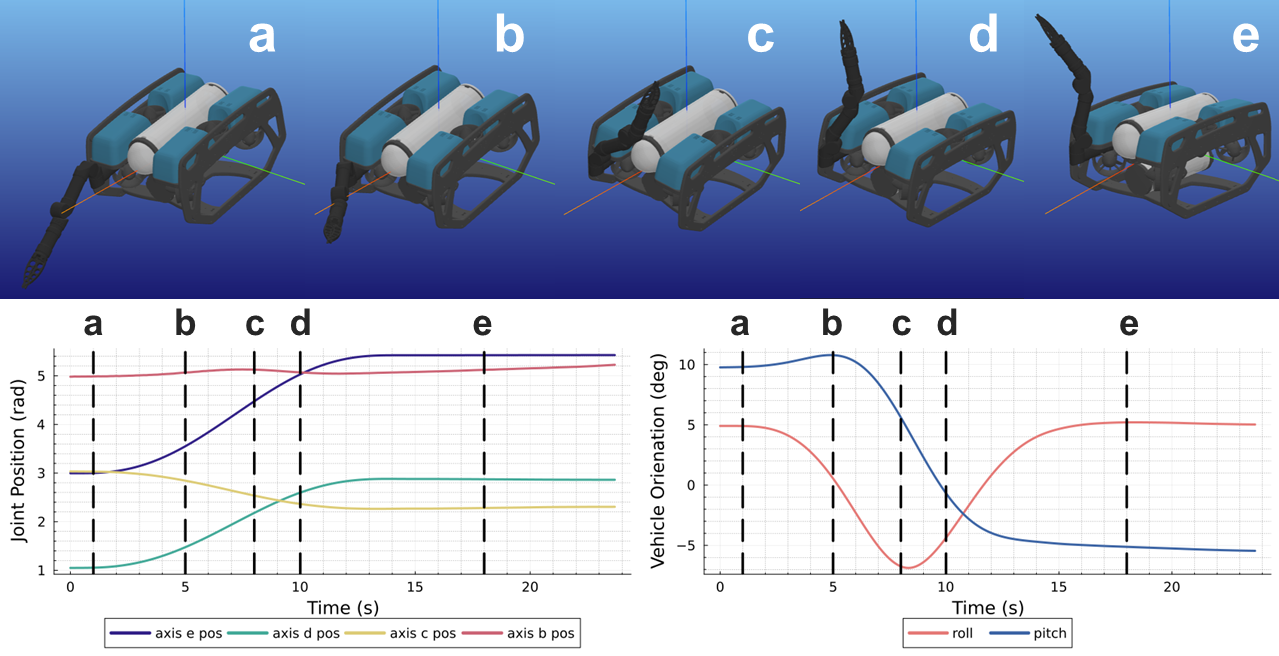}
\caption{\textit{(Top)} Snapshots of the MeshCat visualization of the UVMS performing an example trajectory. \textit{(Bottom left)} Manipulator joint angles over time. \textit{(Bottom right)} Roll, pitch, and yaw of the vehicle over time.}
\label{fig:traj_ex}
\vspace{-10pt}
\end{figure*}

\color{black}

\subsection{Simulation Environment}
\label{sec:sim}

The system comprises a Blue Robotics (Torrance, CA, USA) BlueROV2 vehicle and a Reach Robotics (Sydney, Australia) Alpha 5 electric manipulator \cite{Reach}. The standard version of the vehicle has 4 actuated degrees of freedom (surge, sway, heave, and yaw) and two unactuated DOF which are passively stable (pitch and roll). The Alpha has 4 rotational degrees of freedom and a gripper end effector. Figure \ref{fig:UVMS} is a visual representation of the system and its degrees of freedom. We used the same values for mass, body geometry, buoyancy, added mass, and drag coefficients as seen in the prevailing literature, summarized in \cite{blueROVparams}. We obtained corresponding values for the manipulator from the manufacturer's documentation. 

The vehicle is subject to hydrodynamic forces as described with Fossen's equation \cite{Fossen1994}:
\begin{equation}
\label{eq:fossen}
    M \mathcal{\dot V} + C(\mathcal{V})\mathcal{V} + D(\mathcal{V})\mathcal{V} + g(\eta) = \tau
\end{equation}

where $\mathcal{V}$ is the spatial velocity vector of the vehicle in the body-fixed frame, $M$ is the inertia matrix, $C(\mathcal{V})$ is the matrix of Coriolis and centripetal terms, $D(\mathcal{V})$ is the damping matrix, $g(\eta)$ is the vector of gravitational forces and moments (where $\eta$ refers to the vehicle's pose in the world frame) as well as buoyancy effects, and $\tau$ is the vector of control inputs and external forces. $M$ and $C(\mathcal{V})$ both include a rigid body term (subscript $RB$) and an added mass term (subscript $A$): $M = M_{RB} + M_{A}$ and $C(\mathcal{V}) = C_{RB}(\mathcal{V}) + C_A(\mathcal{V})$. Drag on the vehicle is approximated as the summation of a linear term and a quadratic term. 

However, once a manipulator is added to the system, Fossen's equation has to be solved for every link in the chain. Of the handful of simulators that do this, the most commonly used is UUV Sim, which is built on Gazebo \cite{Gazebo}. Gazebo uses physics engines that solve for the equation

\begin{equation}
    M_{RB}\mathcal{\dot V} + C_{RB}(\mathcal{V})\mathcal{V} + g_{g}(\eta) = \tau_{g}
\end{equation}

where $g_g(\eta)$ represents gravity forces only. This requires hydrodynamic and hydrostatic forces to be included as external torques, such as the forces from the added mass:

\begin{equation}
    \tau_g = -M_{A}\mathcal{\dot V} - C_{A}(\mathcal{V})\mathcal{V} - D(\mathcal{V})\mathcal{V} - g_b(\eta) 
\end{equation}

where $g_b(\eta)$ represents the buoyancy forces. Because the current acceleration $ \mathcal{\dot V}$ is being solved for, only the previous acceleration is available for the right hand side of the equation \cite{UUVSim}. Depending on the value of the added mass, this can cause numerical instabilities in simulation. 

Instead, we adopt the RigidBodyDynamics.jl package in Julia \cite{rigidbodydynamicsjl}. Although Julia's intended use cases reside in the data science realm, its multiple dispatch system makes it extremely fast, a significant benefit for a simulator \cite{Julia}. Additionally, it is entirely open source, providing more transparency into the physics adopted in RigidBodyDynamics.jl. As an alternative method of achieving Equation \ref{eq:fossen}, we set the environment's gravity to zero, and let the inertia of the rigid body equal the sum of the rigid body inertia by itself and the added mass, thereby avoiding the instability issues seen in other solvers. Gravity, buoyancy, and drag are imposed as external forces on each body as usual. The dynamics are solved as if for an open chain of kinematic bodies, with the first joint representing the vehicle a ``floating" joint that imposes no constraints. RigidBodyDynamics.jl is based on Featherstone's implementation of rigid body dynamics \cite{Featherstone_2008} which uses spatial algebra, so external forces can easily be added as external wrenches on the system. Buoyancy and drag are imposed at the geometrical center of each body, whereas gravity is imposed at the center of mass. 

It should also be noted that, since added mass is dependent on geometry, it has direction-dependent values. In our case, the vehicle should have more added mass in the Z direction, since it has more surface area. However, the RigidBodyDynamics package does not support different inertias in different directions, so we simplified the added mass matrix to one mass (the rigid body mass plus the average of the diagonal added mass terms) and a rotational inertia. 

The external forces noted in Equation \ref{eq:fossen} are applied to each link, except linear drag, which is not included for manipulator links. Given the slow velocities seen in simulation and that would be common for manipulation tasks, we believe linear drag would have a negligible effect. 

\subsection{Trajectory Generation}
\label{sec:traj_gen}

In this work, we are interested in how manipulator motion affects vehicle motion. Therefore, we commanded the manipulator to follow predetermined trajectories with no thruster inputs and observed the resulting motions. Each trajectory starts with the manipulator at a random pose in its reachable space. An end waypoint is generated, and a quintic time scaling in joint space is used to interpolate between them. The trajectories ignore self collisions, but respect joint limits on position and velocity. Additionally, a random time scaling between 1x and 2x is applied, such that the duration of each trajectory is variable. An example trajectory and its effect on vehicle orientation in simulation is shown in Figure \ref{fig:traj_ex}.

\subsection{Controller}
\label{sec:control}

The simulation runs at a frequency of 1kHz with a 4th order Runge-Kutta integration. The system state, including the positions and velocities of each joint, is calculated at each time step. Vehicle poses are reported w.r.t. the world frame; vehicle velocities and accelerations are given in the body frame. 

Reach Robotics has proprietary embedded controllers that we approximate with a PID controller. Since the Julia simulation is torque-based, we use a torque PID controller to follow the desired trajectories of the manipulator in simulation. The vehicle is not yet actuated in the simulation, so a controller was not implemented on its actuated degrees of freedom (thruster dynamics are also not modeled). The manipulator controller runs at 100Hz. The torque applied at each joint is calculated as:

\begin{equation}
    \tau_{n+1} = 0.75\tau_n + 0.25\tau_{ff} + \tau_{PID}
\end{equation}
where $\tau_n$ is the torque applied at the previous time step and $\tau_{PID}$ is the PID feedback term, with values as listed in Table \ref{tab:pid}. $\tau_{ff}$ is a feedforward torque comprising the bias term of the system's dynamics equation (the left hand side of Equation \ref{eq:fossen}, minus the inertia term) calculated with the actual simulation system state. The controller updates at a rate of 100Hz. In practice, the arm's embedded controller followed commanded trajectories better than the simulation-based controller. 

\begin{table}
\vspace{5pt}
\caption{PID Controller Gains}
\centering
\begin{tabular}{l l l l}
\toprule
    DoF & \textbf{$K_p$} & \textbf{$K_i$} & \textbf{$K_d$}\\
    \midrule
    Joint E & 3.38e-2 & 4.39e-1 & 1.74e-3 \\
    Joint D & 4.59e-2 & 3.73e-1 & 3.82e-3 \\
    Joint C & 1.35e-2 & 9.64e-2 & 1.27e-3 \\
    Joint B & 9.45e-4 & 1.48e-2 & 2.27e-5 \\
    \bottomrule
    
\end{tabular}
    \label{tab:pid}
    \vspace{-15pt}
    \end{table}

\begin{figure*}
	\centering
\includegraphics[width =\linewidth]{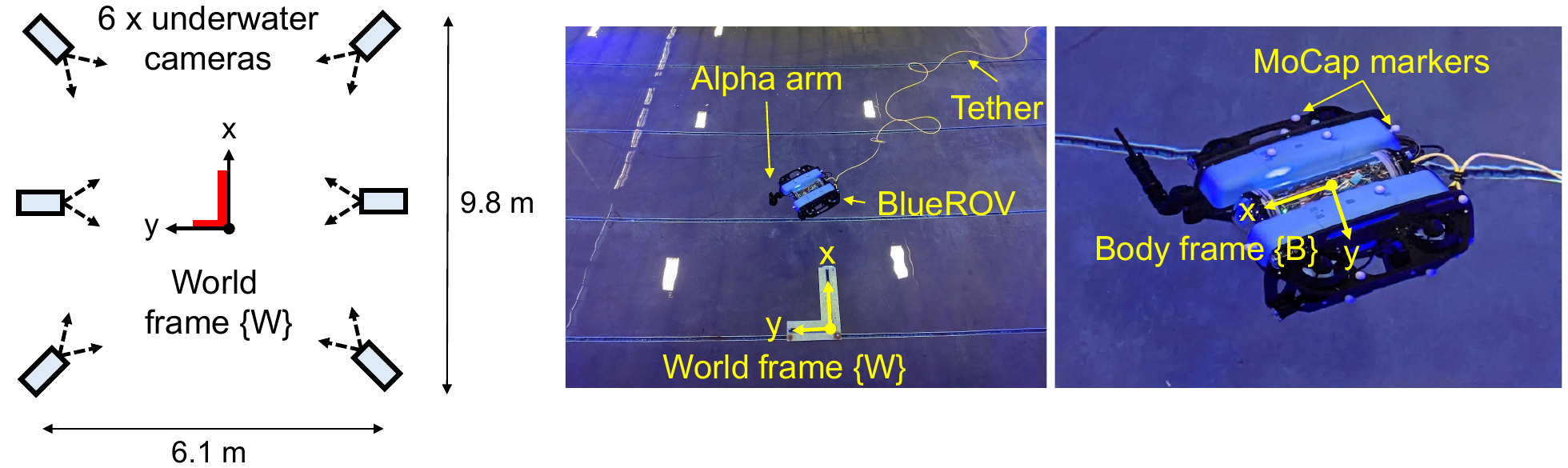}
\caption{\textit{Left} A diagram of the underwater Motion Capture System cameras as placed in the pool. \textit{Center} An image of the BlueROV2 in the pool during experimental trials. \textit{Right} The body frame of the BlueROV2, using a forward - left - up right handed frame. The MoCap markers used to localize the rigid body are in an "L" shape on the top of the vehicle.}
\label{fig:poolsetup}
\vspace{-15pt}
\end{figure*}

\section{Experimental Setup}
\label{sec:experiments}
For physical validation of the model, we completed experimental trials at the O.H. Hinsdale Wave Research Laboratory, placing the physical UVMS system into the wave basin and executing 50 different trajectories. 

\subsection{Motion Capture}
\label{sec:mocap}
We used a Qualisys (G{\"o}teborg, Sweden) underwater motion capture system to measure the UVMS state during experiments. Six cameras were installed in the pool as shown in Figure \ref{fig:poolsetup}. Each camera was fixed about 0.75m from the bottom of the pool, and angled towards the center of the rectangle. Four reflective markers were bolted to the bottom of the pool to indicate the axes of the world frame. Twelve markers were affixed to the outside of the vehicle in an assymetrical pattern. Three of the markers were placed in an `L' shape on the top of the vehicle, as shown in the right image in Figure \ref{fig:poolsetup}, such that the vehicle's x axis and y axis could be set parallel to the line segments between them. The Qualisys MoCap software provided an estimate of the vehicle's pose with respect to the world frame. 

\subsection{UVMS}
\label{sec:UVMS}
The BlueROV was equipped with a WaterLinked DVL A50 (Trondheim, Norway), from which we recorded angular velocity and its onboard, filtered orientation estimate. The Alpha manipulator was connected to a ROS2 driver running a joint trajectory controller via ROS2\_Control. We collected both joint positions and velocities from the manipulator during each trial.

To offset the weight of the arm, four of the provided BlueROV2 weights were attached to the aft side of the vehicle; custom XPS foam was used on either side of the electronics tube in place of the provided foam for slightly more buoyancy. With these additions, the DVL mounted to the back, and the manipulator at its home position, the vehicle was approximately balanced, with a roll of about 4 degrees and a pitch of about 1.5 degrees, as measured by the motion capture. The vehicle was slightly positively buoyant, so it would float to the top of the 1.2-meter water level from the bottom in about 40 seconds.

\subsection{Trial Protocol}
\label{sec:trials}
For each trial, we manually drove the BlueROV to the bottom of the pool with a joystick controller, gave it a few seconds to settle, started recording, and sent a trajectory to the arm. The trajectory executed while the vehicle ascended. We executed 50 different randomly generated trajectories on the system.

\section{Results \& Discussion}
\label{sec:Results}

Before running any trajectories, we collected 3 data points whereby the vehicle simply ascended to the pool's surface over time due to its buoyancy. From the motion capture data, we identified the terminal velocity of the vehicle, which we used to fine-tune the buoyancy of the vehicle in the Julia simulation. We also used these data points to identify the resting pitch and roll of the vehicle with the manipulator at the home position, which were 1.5 and 3.8 degrees, respectively. 

In addition, we discovered that there are slight currents in the pool, possibly from small leaks around the concrete basin, that cause the vehicle to drift: the vehicle moves 8mm/s in the world X direction and -16mm/s in the world Y direction. We did not apply these currents in simulation, but instead gave the simulated vehicle these starting velocities at the beginning of the manipulator trajectory.

\begin{figure}
	\centering
\includegraphics[width =\columnwidth]{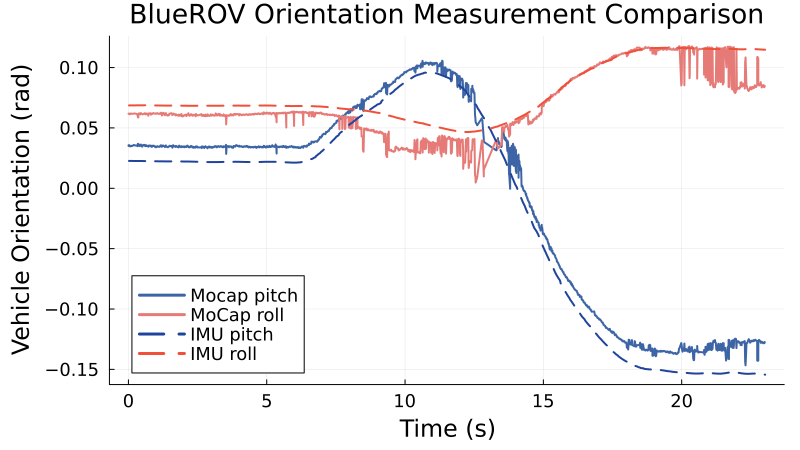}
\caption{Vehicle orientation obtained from the Motion Capture system and the onboard IMU for one trial in which the manipulator followed a trajectory. The MoCap orientation switches frequently between pose estimations based on which of the markers are in view.}
\label{fig:mocapvsimu}
    \vspace{-15pt}
\end{figure}

Both the motion capture (MoCap) system and the onboard DVL measured the orientation of the vehicle during each trial. Figure \ref{fig:mocapvsimu} shows the output from these two sensors during one trial with a manipulator trajectory. Due to the placement of the MoCap cameras, the system was unable to observe enough of the markers to consistently maintain a noise- and bias-free estimate of the vehicle's orientation. In contrast, the IMU data was smooth and quantitatively matched one of the MoCap's estimated poses, so we used IMU data as ground truth to compare against the simulation data. In the future, we will reevaluate camera placements to reduce noisy orientation estimates.

\begin{figure*}[]
\centering
\vspace{5pt}
\includegraphics[scale=0.75]{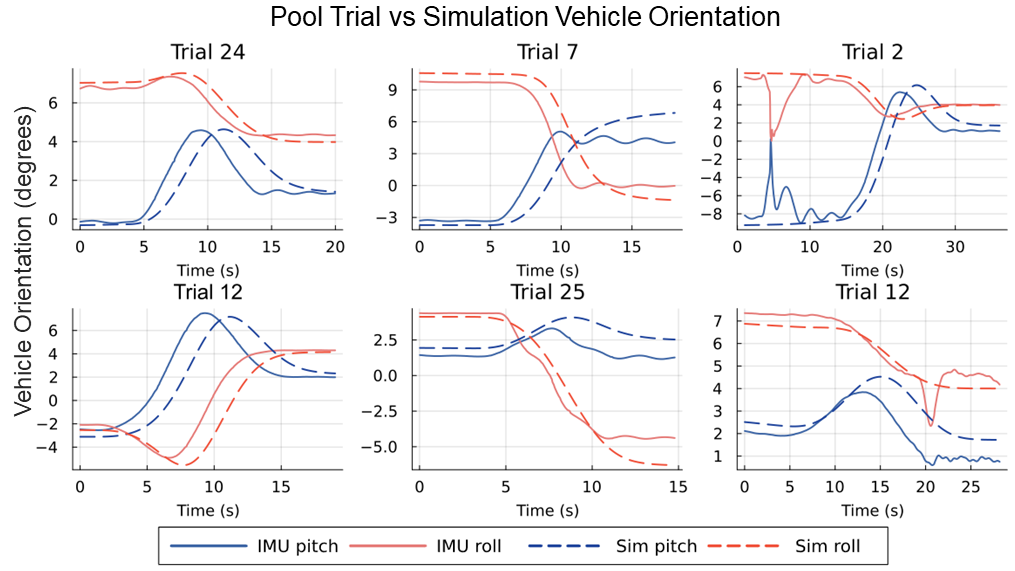}
\caption{A selection of vehicle responses to manipulator motion in simulation and the corresponding experimental trial. \textit{Left column} When the manipulator followed the desired trajectories, we observe good alignment on the intermediate and end orientations of the vehicle, with a slight temporal offset in which the actual vehicle responds faster than the simulated one. \textit{Center column} If the physical manipulator hit a joint limit imposed by the embedded controller, it induced oscillations in the orientation of the vehicle. \textit{Right top} The vehicle hits the bottom of the pool in Trial 2. \textit{Right bottom} The vehicle is actuated in the z direction during this trial, causing a sharp change in roll.}
\label{fig:trials}
\vspace{-10pt}
\end{figure*}

After completing the pool trials, we replicated the manipulator trajectories in the Julia simulation. We set the initial conditions of the simulation to be the same conditions as the pool trials: i) the initial pose and velocity of the vehicle obtained from the MoCap and IMU; and ii) the joint states and velocities from the manipulator's encoders. The simulation was given 10 seconds to settle into its new orientation, and then the trajectory was executed. The simulation data and experimental data were aligned in time by manually matching the joint trajectory sequences. The simulation roll and pitch were then compared to the experimental roll and pitch. The average root mean square error over all 50 trajectories is reported in Table \ref{tab:simerrors}. 

\begin{table}[b]
\caption{Summary Statistics}
\centering
\begin{tabular}{l l l}
\toprule
     & RMSE & StDev\\
    \midrule
    Roll & 1.27 & 0.69\\
    Pitch & 1.73 & 0.82\\
\bottomrule
\end{tabular}
\vspace{4mm}
    \label{tab:simerrors}
    \end{table}
    
A sample of the resulting vehicle motions are plotted in Figure \ref{fig:trials}. In general, when the arm trajectories between simulation and experiment matched, the shape of the vehicle trajectories are qualitatively similar, but have some discrepancies. Primarily, the vehicle responds significantly faster than the model predicts. Most significantly, we found that a 1-2 second temporal 'offset' between the two was consistently present in all trials. 

To further investigate this observation, we calculated the same error statistics if the simulated orientation were to occur 1.6 seconds earlier. With this temporal realignment, the roll and pitch RMSEs drop to 0.86 and 1.03 degrees, respectively -- a 40\% drop in error for pitch and a 32\% drop in error for roll. The consistency of this error indicates that the model is failing to capture some nontrivial portion of the dynamics. Some potential areas of error are: 

\begin{itemize}
    \item \textit{Incorrect hydrodynamic parameters.} Several hydrodynamic terms, such as the added mass and drag terms, have been commonly utilized in the literature, but not necessarily verified. Some of these might be incorrect. 
    \item
    \textit{Modeling error.} It is possible that the added mass simplification we made is unrealistic. Alternatively, some aspect of the system is not adequately captured by the systems of equations we are using to calculate the dynamics. One possible omission could be unmodeled actuator dynamics of the manipulator's motors (e.g. reflected inertia, friction, and backlash). 
    \item \textit{Stochasticity of the energetic environment.} Even when the manipulator is at the home position and the vehicle is unactuated, it settles to slightly different orientations each trial. Across 14 trajectories that started in the manipulator's home position, the standard deviation of pitch and roll were 0.40 and 0.22 degrees, respectively, indicating some degree of randomness that is difficult to model.
\end{itemize}

On a trial-by-trial basis, the degree of alignment between simulation and experiment varied, as shown in Figure~\ref{fig:trials}. The first column represents the set of trajectories that were aligned well between the simulation and the empirical results. The second column represents trajectories that had oscillations during or after the trajectory was executed, which occurred in about 20\% of the trajectories. The beginning of the oscillations matches with a joint suddenly stopping or speeding up. If followed, the trajectories should in theory avoid sudden changes in acceleration due to the quintic scaling, but deviations from this pattern can result in sudden jerks to the system. There is also a setting in the manufacturer's software to prevent the manipulator from colliding with the vehicle, and we suspect some of the requested trajectories violated this collision box, prompting a sudden change in manipulator velocity. This situation is unmodeled in our simulator, but would be an interesting phenomenon to explore further. 

The third column represents edge cases in which external forces were involved. In Trial 2, the vehicle hit the bottom of the pool during the recording, forcing both pitch and roll to 0 degrees momentarily before lifting off again. The ensuing oscillations lasted for 10-20 seconds before damping out. Finally, in Trial 12, we accidentally hit the joystick to move the vehicle downwards during recording, and there is a sharp dip in roll at the same time.

\color{black}
\section{Conclusion}
\label{sec:Conclusion}
Our work introduces a hydrodynamic simulator in Julia for underwater vehicle-manipulator systems specifically designed to investigate the coupling forces between the manipulator and the vehicle. We also demonstrate through pool trials that the model used captures the qualitative shape of the vehicle response and converges to the correct steady state values, but needs some fine tuning to match the speed of the response. We hypothesize that the addition of manipulator actuator dynamics may alleviate some of the discrepancies, and will be explored in future work.

Both our Julia simulator and Stonefish Simulator rely on back end dynamics solvers that use Featherstone’s formulation to calculate the dynamics of kinematic chains, and have integrated hydrodynamic forces on top of that physics engine. Because of this, both utilize a simplification to the added mass model to interface with the underlying rigid body dynamics solvers. Based on the results of our experimental trials, we can say that this is an appropriate simplification that preserves the fidelity of the dynamics, and the output can be relied upon to be accurate within a couple degrees.



\bibliographystyle{./bibliography/IEEEtran}
\bibliography{./bibliography/PitchRNN2022.bib}

\end{document}